\documentclass[letterpaper,twocolumn,10pt]{article}
\usepackage{usenix2019_v3}

\usepackage{tikz}
\usepackage{amsmath}

\usepackage{graphicx}
\usepackage{amssymb}
\usepackage{booktabs}

\usepackage{times}
\usepackage{epsfig}
\usepackage{tabularx}
\usepackage{subcaption}
\newcolumntype{b}{X}
\newcolumntype{s}{>{\hsize=.5\hsize}X}

\usepackage[backend=biber,style=ieee,]{biblatex}
\begin{document}

\date{}

\title{\Large \bf Using Color To Identify Insider Threats}

\author{{\rm Sameer Khanna}\\ Department of Computer Science, Stanford University\\ Department of Research and Development, Fortinet\\}

\maketitle

\begin{abstract}
Insider threats are costly, hard to detect, and unfortunately rising in occurrence. Seeking to improve detection of such threats, we develop novel techniques to enable us to extract powerful features and augment attack vectors for greater classification power. Most importantly, we generate high quality color image encodings of user behavior that do not have the downsides of traditional greyscale image encodings. Combined, they form Computer Vision User and Entity Behavior Analytics, a detection system designed from the ground up to improve upon advancements in academia and mitigate the issues that prevent the usage of advanced models in industry. The proposed system beats state-of-art methods used in academia and as well as in industry on a gold standard benchmarking dataset.
\end{abstract}

\section{Introduction \label{section:intro}}

As we move further into an ever more digital age, newer and more complex attack vectors are appearing everyday. Insiders pose a unique threat to corporations and organizations of all scales due to their access to proprietary systems and their ability to circumvent security protocols and blind spots the public is not privy to. Close to 30\% of confirmed breaches today involve insiders \cite{verizon2018data}. In total, over 2,560 internal security breaches occur in United States businesses every day \cite{IsDecisionThreatManifesto} with a year-over-year increase in insider attack rates of 21.4\% \cite{solutions2019data}. Each such attack costs an organization on average 11.45 million USD annually \cite{PonemonCostThreat}.

Unfortunately, these attacks are extremely difficult to detect. Third party entities detect the vast majority of most data breaches that occur within an organization long after the damage has been done; famous examples include the breaches in TJX Companies, VeriSign, Adobe and LinkedIn \cite{TJXSecurityBreach,VeriSignBreach,AdobeBreach,LinkedInBreach}. 

The most popular method of approach to this problem is framing insider threat detection as an anomaly detection problem \cite{berman2019survey}. While these approaches benefit greatly by being able to detect a variety of different attack patterns, these models tend to have poor precision when compared to supervised learning counterparts \cite{chandola2009anomaly}. A poor precision indicates the detection model is labeling many employees that have done nothing wrong as planning internal attacks, something that is not acceptable when livelihoods can be at stake.

A recent trend in the cybersecurity space to achieve greater accuracy in detection of threats is to encode information via greyscale images. These encodings enable the usage of natural image transfer learning in order to fine-tune models trained on large high-quality datasets like ImageNet \cite{deng2009imagenet} for great predictive power on downstream tasks without needing large amounts of data related to the task in question \cite{he2016deep, rezende2017malicious, kancherla2013image, tobiyama2016malware, rumelhart1985learning, lecun1999object}. Indeed, such implementations have become so prevalent in cybersecurity that its ubiquity has been highlighted in malware detection, phishing detection, and network traffic anomaly detection \cite{zhao2020review}. This natural image transfer learning approach has been proposed to be used for insider threat detection as well, leading to state of the art detection performance \cite{gayathri2020image}.

The main issue with forcing non-spatial data into a spatial format via such greyscale image encodings is that the convolutional prior would be inducing a fundamentally untrue prior into the computer vision model architecture.  Recent in-depth analysis on this approach in regards to its application to malware detection has show that it leads to subpar models that are brittle and fail to generalize; the accuracy improvements reported by such techniques were discovered to be the result of over-fitting on bad data \cite{raff2018malware}.

We seek to improve upon the shortcomings of recent research into the use of image encoding approaches in cybersecurity research by converting the insider threat detection problem into the simpler problem of detecting color in an image. Our approach utilizes a novel image encoding paradigm where malicious behavior encodings are colorful whereas benign behavior encodings are more muted and grey in appearance. This approach to image encodings sidesteps the convolution prior issue altogether and allows one to formulate an unsupervised anomaly detection problem into a supervised binary classification problem, leading to models that generalize well to unseen data and zero-day attack vectors with state of the art performance. We believe this approach is highly applicable to a variety of cybersecurity sub-fields like malware detection and phishing detection, as well as anomaly detection as a whole.

\textbf{Contributions:  } In addition to the novel image encoding paradigm, we propose some additional mechanisms in order to improve insider threat detection performance. Our contributions are as follows: 

\begin{itemize}
\item We define File Path Variance, a metric that measures the degree a user searches an organization files.
\item We utilize Natural Language Processing techniques to extract novel features from email and website access data such as indications of a disgruntled employee.
\item We represent behavior indicators as novel image representations designed such that color is an indicator of malicious behavior, allowing identification of behavioral changes at a glance.
\item We propose a novel context changing data augmentation that addresses data imbalance issues while preserving the unique composition of image encodings.
\item We use a dual input classifier architecture, feeding in non-dynamic information to reduce the number of false positive behavior classifications.
\end{itemize}

Together, these allow our proposed model, Computer Vision User Entity Behavior Analytics (CVUEBA), to outperform state-of-the-art insider threat models on a gold standard benchmarking dataset.

\section{Related Works \label{section:relatedworks}}

\textbf{Academia:  } Researchers have worked on a plethora of solutions for insider threat detection, the vast majority of which utilize machine learning. 

Gavai et al. \cite{gavai2015detecting} proposed an Isolation Forest-based unsupervised approach for detecting insider threats using network logs. They aggregate features that contribute most to the isolation of a data sample within the tree to better ascertain why a user was tagged as anomalous.

Liu et al. \cite{liu2018anomaly} proposed an ensemble of deep autoencoders to unravel the non-linear relationships in log data. A model is built from each autoencoder based on the extracted features from each log file. Finally, the outputs are aggregated into a single model used to detect malicious insider activities. Unfortunately this procedure has its limitations: returning subpar results for datasets from alternative sources, the frequency based feature extraction methodology does not always provide the expected outcome, and the one-hour interval considered for user behavior study does not provide enough resolution to identify usage patterns.

Noever et al. \cite{noever2019classifier} tried a variety of different learning algorithm families, concluding that Random Forests with Randomization and Boosted Logistic Regression provided the best results after extracting risk factors from data to create their feature vectors. While their results indicate that Boosted Logistic Regression outperformed the former algorithm, Noever et al. advocate for the usage of Random Forests in insider threat detection systems as they offer a deep but human-readable set of detection rules.

Noting that the vast majority of implementations suggested in recent publications suffer from a very low accuracy of the minority class due to extreme class imbalance, Al-Mhiqani et al. \cite{al2021integrated} proposed an intuitive way to tackle this issue. They combine adaptive synthetic sampling (AD) \cite{he2008adaptive} with a deep neural network (DNN) architecture to develop AD-DNN, an integrated model that improves the overall detection performance of insider threats. While the paper produces promising results, AD has traditionally led to brittle models that fail to generalize well to unseen data \cite{gosain2017handling}.

Sharma et al. \cite{sharma2020user} used a two-step process to detection via their Long Short-Term Memory Autoencoder (LSTM-Autoencoder). First, it calculates the reconstruction error using the autoencoder fit on normal data, and then utilizes a threshold based detection scheme to identify outliers. The identified outliers are then classified as malicious behavior.

Le et al. \cite{le2021training} assessed various semi-supervised learning algorithms in conjunction with different labeled data availability conditions. These conditions are designed to emulate real-world situations representing the availability of various scenarios of ground truth.

Meng et al. \cite{meng2018deep} combined Long Short-Term Memory Recurrent Neural Network (LSTM-RNN) and Kernel Principal Component Analysis (PCA) for analysis of insider behavior. They compared well against popular algorithms such as Support Vector Machines (SVM) and Isolation Forests, however it is important to note that their approach was not compared with deep learning models.

Yuan et al. \cite{yuan2018insider} identified that a user action sequence has a temporal dependency. They feed these sequences to a Long Short-Term Memory (LSTM) network, which extracts user behavior features and predicts the next user action. Various hidden states of the LSTM are then used to develop a fixed size representation passed to a Convolutional Neural Network (CNN) for detection purposes.

It is beneficial to identify user behavior patterns within multi-domain scenarios. However, incorporating multi-domain irrelevant features may hide the existence of anomalies within our data. Thus, Lin et al. \cite{lin2017insider} formulated a hybrid method using Deep Belief Networks (DBN) for unsupervised feature reconstruction, and One Class SVM (OCSVM) for insider threat detection. The usage of the DBN provides a substantial performance uplift when compared to using OCSVM by itself, indicating a promising direction for insider threat detection. 

Lin et al. are not the only team that proposed using this network family; Zhang et al. \cite{zhang2018insider} also focused on using a DBN network, albeit within a supervised regime. First, feature learning is executed by having each layer trained using the unsupervised learning method and the training results are adopted as the input of the next layer. Finally, the entire network is fine-tuned by using supervised training. The final output is determined after being fed through a softmax output layer.

Chattopadhyay et al. \cite{chattopadhyay2018scenario} proposed an insider threat detection approach based on classification of time-series user activities. A cost-sensitive technique for data adjustment was used to randomly undersample the instances belonging to the minority class. A deep autoencoder with two layers and a threshold parameter was used for classification.

Despite academia on the topic of insider detection dating as far back as 1987 with Denning's anomaly detection regime \cite{denning1987intrusion}, very little has reached the industry. Rieck et al. \cite{rieck2011computer} cites reasons why current research in academia is not well-aligned with industry requirements such as the high cost of false positives, the semantic gap between results and their operational interpretation, and difficulty in performing sound evaluation. These requirements are sometimes in direct opposition to assumptions made in academia; for example, academic insider threat systems frequently seek to improve recall at the cost of precision \cite{yuan2021deep}.

\textbf{Industry:  } Thus, while numerous User and Entity Behavior Analytics systems (UEBA), systems built in the security industry to detect insider threats, have been created, most rely on their own datasets and experience rather than using discoveries found in academia. 

There are few vendors that have publicly detailed how their UEBA systems operate. Examples of companies providing industry implementations include Niara, which utilizes Mahalanobis distance outlier detection \cite{shashanka2016user}, Fortinet, which uses Naive Bayes to categorize activity by an anomaly score \cite{fortiinsight}, Exabeam who uses a second order Factorization Machine to improve first-time access malicious activity reporting \cite{tang2017reducing}, and Aruba Networks who proposes models using Support Vector Machines, and Logistic Regression \cite{arubacisoguide}.

Unfortunately, while there are numerous UEBA systems to be found in industry, most are rudimentary in nature. Bussa et al. in Gartner's market analysis report for UEBA states that most vendors still rely at least partially on rule based implementations and require upwards of half a year worth of tuning in order to achieve effectiveness \cite{bussa2018market}.

\section{Behavior Encodings}

Insider threat systems rely on a large variety of data sources from  many different resources including login data, LDAP information, website and file access, external drive data, and email activity. Thus, it is crucial to obtain feature representations of a user's behavior in order to appropriately utilize vital information. While we glean insights from previous work in order to define our representations \cite{chattopadhyay2018scenario, gayathri2020image, kim2019insider}, we also propose novel features which we detail in the following sections. Through careful review, analysis and feature engineering, we arrive at the set of features detailed in Table \ref{table:features} alongside the data source each feature is derived from.

\begin{table}[htbp]
\vspace{-4mm}
\footnotesize
\caption{\label{table:features} Set of features compiled to assess user behavior}
\begin{tabularx}{\linewidth}{c*{1}{>{\raggedright\arraybackslash}X}}
\toprule
Data Source & \multicolumn{1}{c}{Feature}                                                                      \\\toprule
LDAP          & User's ID                                                                                    \\ \midrule
All Sources   & Date                                                                                         \\ \midrule
Logon         & Difference between initial logon and office start time.                                      \\ \midrule
Logon         & Difference between last logon and office start time.                                         \\ \midrule
Logon & Average difference in time between office start time and number of logins before office hours. \\ \midrule
Logon    & Average difference in time between office end time and number of logins after office hours.  \\ \midrule
Logon         & Total number of logins.                                                                      \\ \midrule
Logon         & Total number of logins outside office hours.                                                 \\ \midrule
Logon         & Total number of logoffs.                                                                     \\ \midrule
Logon         & Total number of logoffs outside office hours.                                                \\ \midrule
Logon         & Total number of unique systems accessed.                                                     \\ \midrule
Logon         & Total number of unique systems accessed outside office hours.                                \\ \midrule
Logon         & Average session length held outside office hours.                                            \\ \midrule
Device        & Total number of external device usages.                                                      \\ \midrule
Device        & Total number of external device usages outside office hours.                                 \\ \midrule
File          & Number of executable files downloaded, run, or handled in some form.                         \\ \midrule
File          & File Path Variance throughout the day.                                                       \\ \midrule
File          & File Path Variance after office hours.                                                       \\ \midrule
Email         & Number of emails sent outside organization domain.                                           \\ \midrule
Email         & Number of recipients supervisor has sent emails to within organization domain.               \\ \midrule
Email         & Number of attachments sent with emails.                                                      \\ \midrule
Email         & Average size of emails.                                                                      \\ \midrule
Email         & Total number of email recipients.                                                            \\ \midrule
Email         & Number of emails Conical Classification identified as the user being disgruntled.            \\ \midrule
HTTP          & Number of websites Conical Classification identified as being job posting sites.                     \\ \midrule
HTTP          & Number of websites Conical Classification identified as being Wikileaks or Wikileaks clones. \\ \midrule
HTTP          & Number of websites Conical Classification identified as being keylogger download sites.      \\
\bottomrule
\vspace{-8mm}
\end{tabularx}
\end{table}

\subsection{Temporal Contextualization of Features}
The vast majority of existing insider threat approaches focus on activity type information, such as copying files to a remote drive. This has proven to be ineffective, as users can perform the same activity with either benign or malicious intent \cite{yuan2021deep}. Temporal contextualization plays an important role in analyzing user behavior to identify malicious intent; while copying files to a drive during working hours can be considered normal behavior, the same behavior in the middle of the night should be considered malicious in nature as such activity likely indicates the user is stealing intellectual property, or the user is installing malware or a keylogger onto a system. 

Typical office hours are between 9am and 5pm \cite{officehoursworld}, however we define office hours as being from 7am to 7pm, a two hour buffer on each end, in order to improve our identification of malicious behaviors. A user entering the office a bit earlier or leaving the office a bit later than their peers should not be considered as potentially malicious behavior.

\subsection{Quantifying File Access Risk Exposure}

\begin{figure*}[htbp]
\centerline{\includegraphics[width=\linewidth,keepaspectratio]{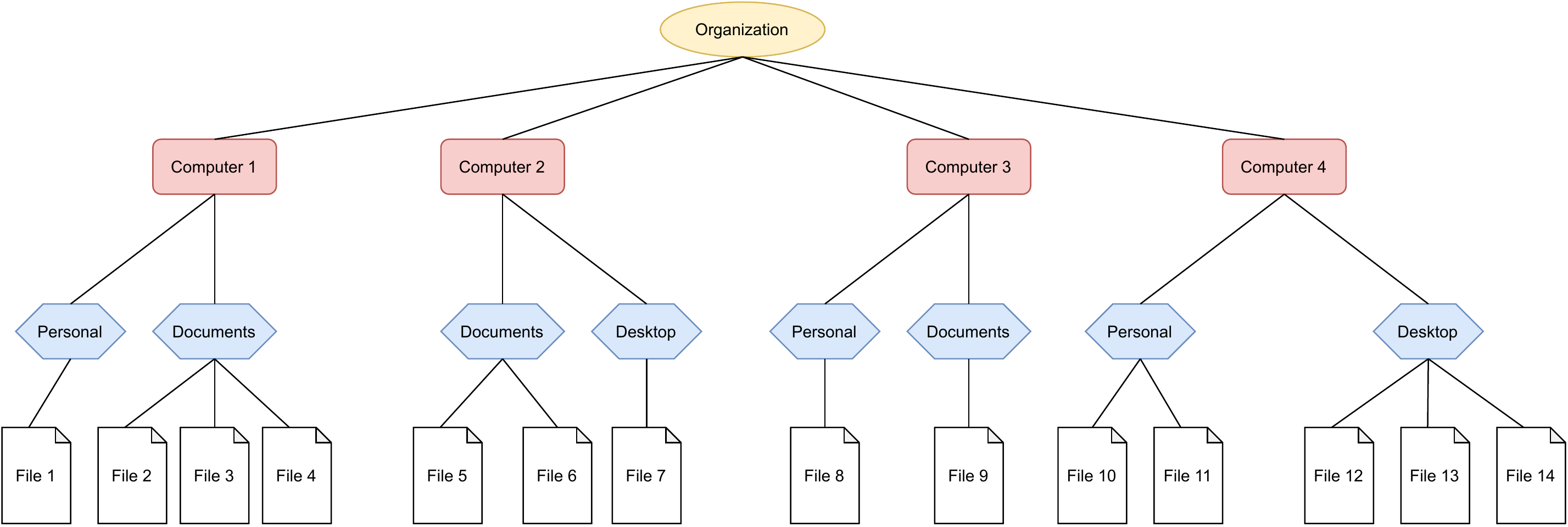}}
\caption{\label{figure:filetree} File tree created for an example organization. The root node is the organization itself, its branches are all computers and servers within the organization, sub-directories are represented via further branching, and files are the terminal leaves within the tree structure. Edge weights, represented by the thickness of edges, increases as the files being accessed move further outside a user's sphere of permissions.}
\label{fig}
\end{figure*}

\textbf{Tree Representation:} A given user accesses files within certain locations for the vast majority of their work, typically on their work computers, shared access computers, or perhaps on their colleagues computers from time to time. 

If a user ventures away from such access behavior, for example rapidly expanding the number of computers they access files on or venturing into directories on shared computers irrelevant to their work or previous interest, such behavior should be flagged for a human expert to delve into and perform risk assessment on. Such behavior is indicative of an employee attempting to search for sensitive, compromising, or confidential information \cite{chan2003corporate, bressler2014protecting, vashisth2013corporate, mashingaidzecorporate, hansen2020spy}.

We represent all the files within an organization on a given day via a tree structure, with leaves representing different files, and branches representing different computers, directories, and sub-directories. The further apart leaves are within the generated structure, the more distance the respective files are from each other within the organization. Figure \ref{figure:filetree} illustrates an example tree structure for accessed files of a user in an organization. 

The distance $d_{ij}$ between file paths $i$ and $j$ is defined as shown in Equation 1, where $lca$ refers to the lowest common ancestor of leaves $i$ and $j$, and $Dist(n1, n2)$ is the distance between nodes $n1$ and $n2$.

\begin{equation}
    \resizebox{0.7\linewidth}{!}{$
    \begin{aligned}
    d_{ij} = Dist(root, i)
    &+ Dist(root, j) \\
    &- 2*Dist(root, lca)
    \end{aligned}
    $}
\end{equation}

\textbf{File Path Variance:  } From this representation, we develop File Path Variance (FPV), whose formula is defined in Equation 2. As indicated by the name, the FPV calculates the variance of the leaves in our file tree structure that have been accessed by a given user. FPV enables us to quantify file access in an easily interpretable manner; a low FPV signifies a user spending their workday working on a small set of files they have the right permissions for, while a high FPV may indicate a user is searching through numerous files and computers for sensitive information.

\begin{equation}
    \resizebox{0.45\linewidth}{!}{%
    $FPV = \frac{\sum_{i=1}^N\sum_{j=1}^N \mathbf{1}\{i \neq j\} d_{ij}^2}{2N^2 - N}$%
    }
\end{equation}

\subsection{Parsing Text Information Sources}
A single user consumes and produces hundreds to thousands of different bodies of text, such as websites and emails, on a daily basis; being able to identify concerning topics and sentiment from this data would prove beneficial in identifying potential threats. Unfortunately, there is a relative lack of content variety discussing how to identify text on a particular subject from a corpus consisting of a variety of subjects in a One-Vs-All configuration, especially with low computational costs. In practice, this leads to sub-par insider threat detection models for the sake of speed. 

For example, despite insider threat detection primarily working with log and textual information, the vast majority of publications do not utilize Natural Language Processing in their implementations \cite{wei2021insider, tuor2017deep, meng2018deep, le2018benchmarking, le2019machine, gayathri2020image}. Many that do follow Chattopadhyay et al.'s approach of simply summing over TF-IDF vectors before feeding the result as a feature into detection models \cite{chattopadhyay2018scenario}.

In order to isolate topics of interest with high computational efficiency, we utilize Conical Classification (CC), a computationally efficient method for one class topic determination \cite{khanna2021conical}. A document is first converted to a vector representation using the Vector Space Model (VSM) Normal Exclusion (NE), whose formula is defined in Equation 3. Here, tpr is the true positive rate $P(word | positive class)$ as determined via $\frac{tp}{pos}$, where $tp$ is the number of positive training cases containing the word and $pos$ is the number of positive training cases. $F^{-1}$ is the inverse  Normal cumulative distribution function. $\epsilon$ is a number with small magnitude added to avoid the undefined scenario of $F^{-1}(0)$; here, we set $\epsilon$ to $0.0005$. 

For our purposes, we compiled the frequencies of the top $\frac{1}{3}$ million words in the human language using Tatman's English word count dataset \cite{tatmankaggleword, brantsgooglewordset} and stored them within a dictionary for rapid lookup. We can safely set the frequencies of words that do not appear in our dictionary to 0, as these include words that rarely appear in standard language; such words include abaptiston, abaxile, grithbreach, gurhofite, zarnich, and zeagonite. Indeed, according to Oxford's compiled statistics, the frequency of occurrence for all such words combined represent approximately a percent of the entire lexicon of the English language, easily within the margin of error for our analysis \cite{oecstats}. We store these compiled frequencies within a dictionary and utilize them within our calculations of the NE, which is indicated within our equation via $Dict[word]$.

\begin{equation}
    \resizebox{0.8\linewidth}{!}{%
    $NE = \left|F^{-1}(tpr + \epsilon) - F^{-1}(Dict[word] + \epsilon)\right|$%
    }
\end{equation}

NE is called as such as it excludes, or reduces, the weightage of words that are inconsequential to determining the topic of text without requiring a negative corpus to be present. We will scale NE by the frequency of the term within the corpus (TF) for our model developing the NE-TF VSM. Our representation of a word in a model will thus be determined by how frequently a word occurs in our corpus, scaled by the statistical significance of the word within the evaluated text. Higher magnitude values give a strong indication that the vector is about our target topic, while lower values would lead to a lower confidence that such a conclusion is correct.

Next, CC determines if the vector representation of the document is within the conical span of the positive training set of vectors. The conical span is defined in Equation 4. Here, $\lambda_1 + ... + \lambda_k$ correspond to non-negative scalar values, and $v_1 + ... + v_k$ refer to the vector representations for all documents in our positive corpus.

\begin{equation}
    \resizebox{0.9\linewidth}{!}{%
    $conical(S) := \left\{\lambda_1v_1 + ... + \lambda_kv_k : \lambda_1 + ... + \lambda_k > 0 \right\}$%
    }
\end{equation}

Designed primarily for use in insider threat detection systems, CC can identify when a user has gone to websites of note such as job forums, keylogger download hubs, and Wikileaks-like sites, as well as if a user is disgruntled with their job as indicated from the emails they have sent. All of these are indicators of a user that is not happy with their current situation, have strong historical correlations with the execution of insider attacks \cite{salem2008survey, slocombe2014beware, stolfo2008insider, pfleeger2008reflections}, and demand additional vigilance into activities to prevent future attack.

\subsection{\label{subsection:greyscale} Greyscale Image Encodings}

\begin{figure*}[!h]
\centering
\begin{subfigure}[b]{\textwidth}
   \includegraphics[width=\linewidth,keepaspectratio]{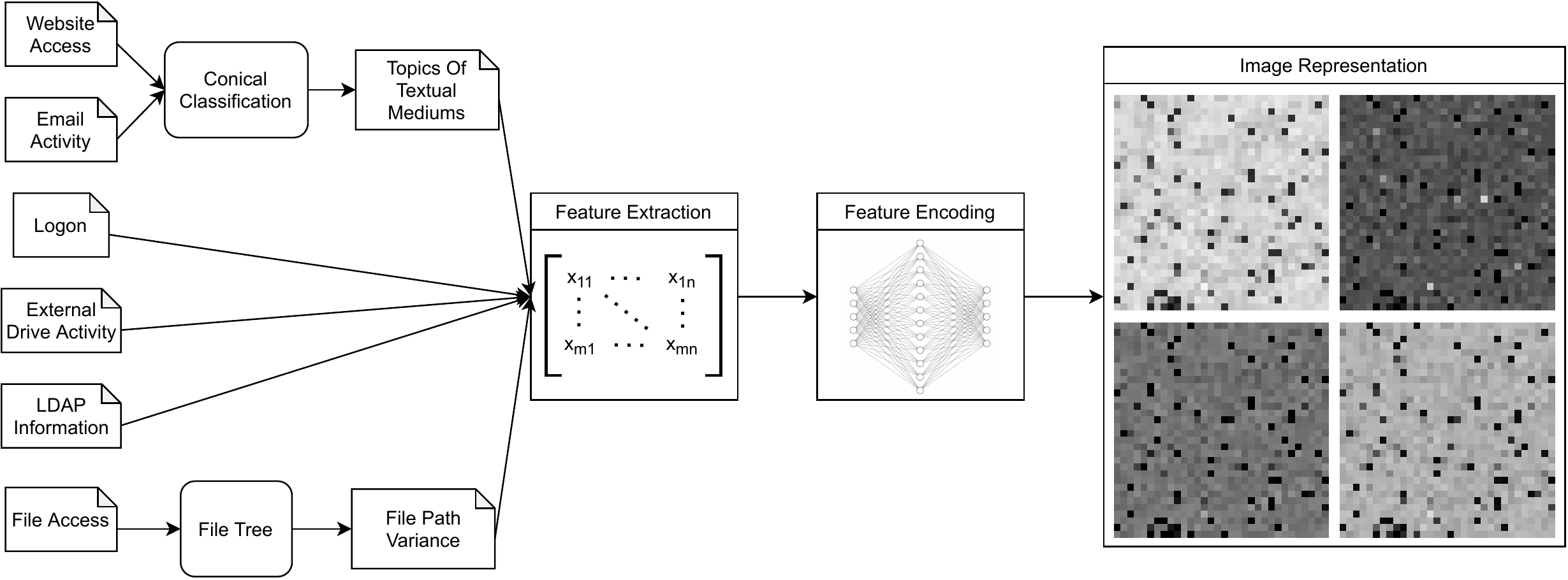}
   \caption{}
   \label{figure:FeatureToGrey}
\end{subfigure}

\begin{subfigure}[b]{\textwidth}
   \includegraphics[width=\linewidth,keepaspectratio]{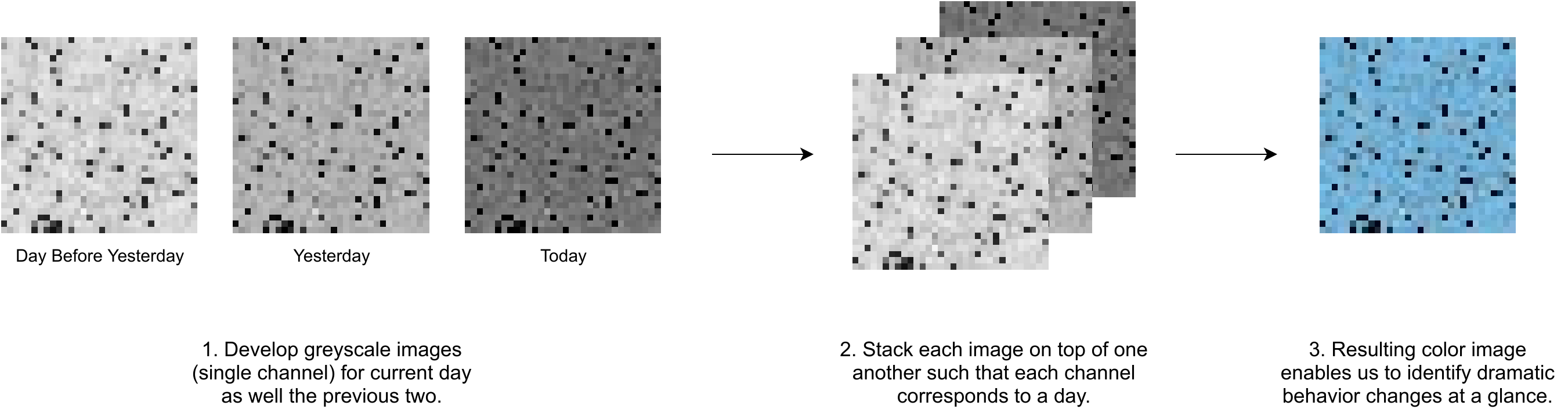}
   \caption{}
   \label{figure:greyToColor}
\end{subfigure}

\caption{Generating Color Image Encodings from start to finish. (a) Process flow from log files to greyscale images. (b) Representing context behavior encodings as channels.}
\end{figure*}

We can now transform the extracted behavior information into image representations. Unlike traditional image encoding implementations that utilizes interpolation \cite{gribbon2004novel} to facilitate this process, we take advantage of Sparse AutoEncoders (SAE).

The loss function for a SAE can be found in Equation 5. $D_{KL}$, defined in Equation 6, stands for the Kullback-Leiber Divergence between a Bernoulli random variable with mean $\rho$ and Bernoulli random variable with mean $\hat{\rho_j}$. $\rho$ is a sparsity hyperparameter of the model that constrains the average activation of each neuron $j$ to be as close to itself in value as possible. $\hat{\rho_j}$, defined in Equation 7, is the average activation of hidden unit $j$.

\begin{equation}
\resizebox{0.55 \linewidth}{!}{%
$L_{sparse} = L + \beta\sum_j D_{KL}(\rho || \hat{\rho_j})$%
}
\end{equation}
\begin{equation}
\resizebox{0.65 \linewidth}{!}{%
$D_{KL}(\rho || \hat{\rho_j}) = \rho\log\frac{\rho}{\hat{\rho_j}} +
(1 - \rho)\log\frac{1 - \rho}{1 - \hat{\rho_j}}$%
}
\end{equation}

\begin{equation}
\resizebox{0.35 \linewidth}{!}{%
    $\hat{\rho_j} = \frac{1}{m}\sum_{i=1}^m a_j(x^{(i)})$%
}
\end{equation}

SAEs in anomaly detection are typically trained on normal data only. The expectation is that the reconstruction error will be noticeably higher on anomalies than it will on normal data, as anomalies will be encoded differently and thus will be distributed away from normal data. A threshold parameter is then used to separate vectors into normal and anomalous classes \cite{chen2018autoencoder}. We instead use the trained SAE hidden layer to automatically learn better feature representations from the given data \cite{ng2011sparse}.

To transform the SAE output values into a range of 0-255 suitable for images, we first apply Min-Max Scaling and proceed to multiply all values by 255. Both of these actions are done within the model's layers itself; our testing has shown that performing the scaling in this manner improves speed, as well as storage and memory requirements for training. Our SAE has a hidden dimension of size 1024; once output encodings are scaled, they are reshaped into 32x32x1 images. Figure \ref{figure:FeatureToGrey} details the full process used to convert text based log data into greyscale images.

\subsection{\label{subsection:contextchannel} Context-Channel Representations}

While we could feed these greyscale images directly into a model for attack identification, it would be beneficial to have contextual information regarding the typical behavior of a user. This would enable us to identify the sudden behavioral changes indicative of an attack and mitigates the concerns regarding insider threat detection systems brought up by Tan et al. \cite{tan2002undermining}; by comparing a user's behavior to their own previous behavior rather than performing the comparison at the activity level via a detection regime, we mitigate the potential of attackers taking advantage of detection loopholes and acting undetected. 

We provide this information by passing in behavior encodings for the previous two days in addition to the current day we are evaluating. As shown in Figure \ref{figure:greyToColor}, we append the encoding for each day as a different channel, leaving us with a color image for evaluation purposes.

If a user is not acting malicious, there is little to no variation in behavior features from a day to day basis; this leaves images representing benign behavior fairly grey in appearance. On the other hand, malicious actions will indeed have changes in observed behavior, which will lead to colorful images. 

This phenomenon is due to how RGB images work. The RGB scale is calibrated so that when a given pixel's red-green-blue numbers are equal, the pixel is represented as a shade of gray where larger red-green-blue numbers lead to lighter shades of grey. As discussed in Section \ref{subsection:greyscale}, our SAE model is trained such that the red-green-blue numbers across all dimensions will be fairly similar to one another if a user's behavior is benign. Malicious behavior on the other hand will lead to drastic differences in the red-green-blue numbers as the current information varies significantly from the contextual information which leads to the image having a more colorful representation. 

\begin{figure}[htbp]
\centerline{\includegraphics[width=\linewidth,keepaspectratio]{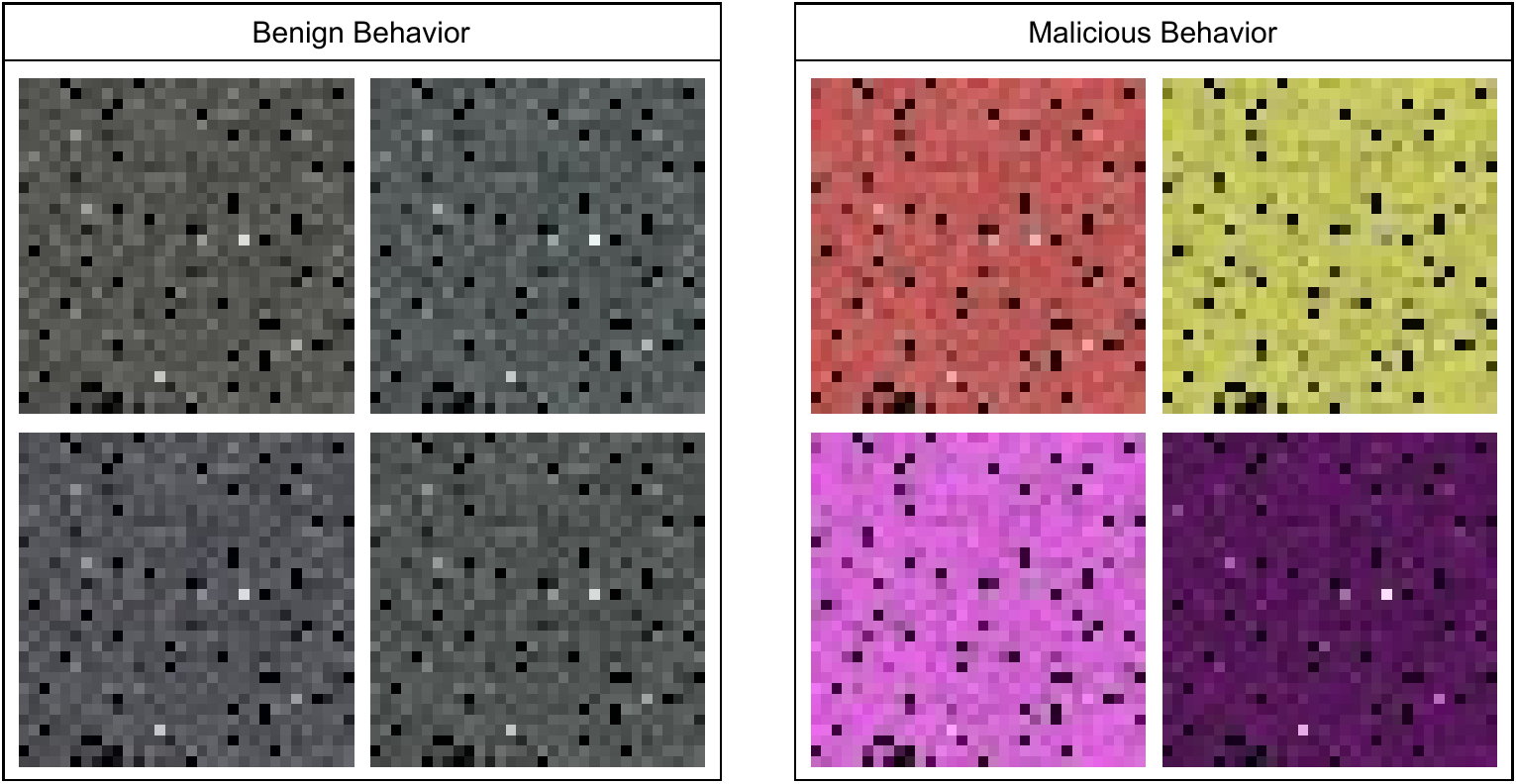}}
\caption{\label{figure:benignVsMalicious} Benign vs Malicious Images.}
\label{fig}
\end{figure}

As we can see in Figure \ref{figure:benignVsMalicious}, this allows malicious image representations to be easily identifiable, even by those who are not security experts. Using Tree Parzen Estimation hyperparameter tuning (TPE) \cite{bergstra2013making}, we maximize the separability of malicious behavior from benign behavior as determined via linear evaluation, keeping the parameter set that achieved the best performance. Thus, when malicious behavior is seen, we see cross-channel differences across the vast majority of dimensions, leading to the malicious images taking similar shades of color throughout the image encoding.

\textbf{Alternative Channel Representations:  } We utilize a daily context channel representation as it allows for maximum granularity in regards to detecting variations in a user's actions. However, it is imperative to take into account that such a representation will not always fit the needs and desires of an organization's security team given the unique layout of the organization's data. After all, evidence suggests that the perception of information technology (IT) behavior in the workplace can be widely varying. In a recent survey of 500 IT leaders and 4000 employees, significant discrepancies were observed in perceptions of insider threats \cite{egresssurvey19}.

To accommodate this, we can replace what information our background encodings represent. For example, if an organization wishes to track user behavior when compared to their historical trends, the first baseline channel could be represented by average feature values from the beginning of a person's time at the organization, while the second baseline channel could be the average feature values from the past week. If instead we wish to compare a user's behavior against a particular role, the first channel could be the average values from all employees in the given role, while the second channel can be the average value for the user's teammates.

\begin{figure*}[!h]
\centering
\begin{subfigure}[b]{\textwidth}
   \includegraphics[width=\linewidth,keepaspectratio]{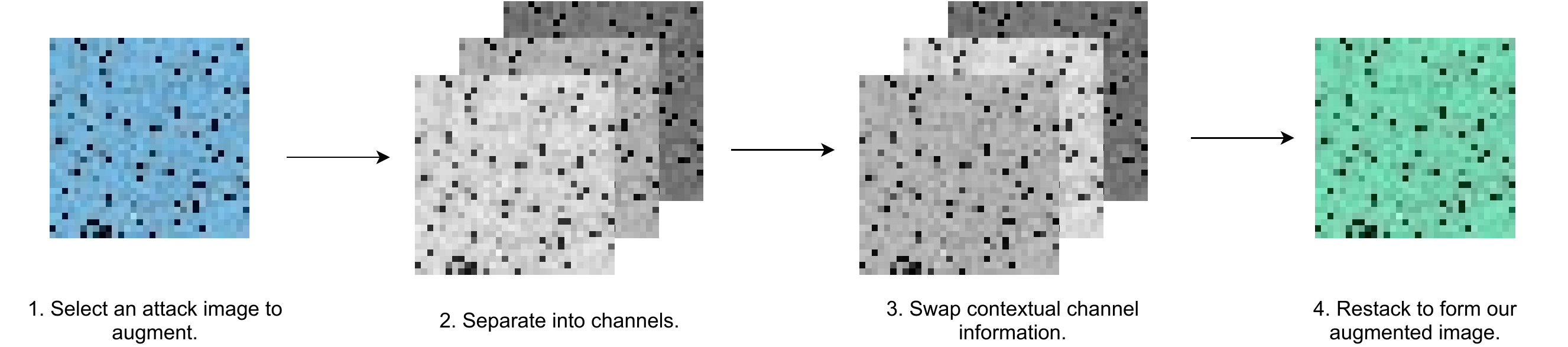}
   \caption{\vspace{2mm}}
\end{subfigure}

\begin{subfigure}[b]{\textwidth}
   \includegraphics[width=\linewidth,keepaspectratio]{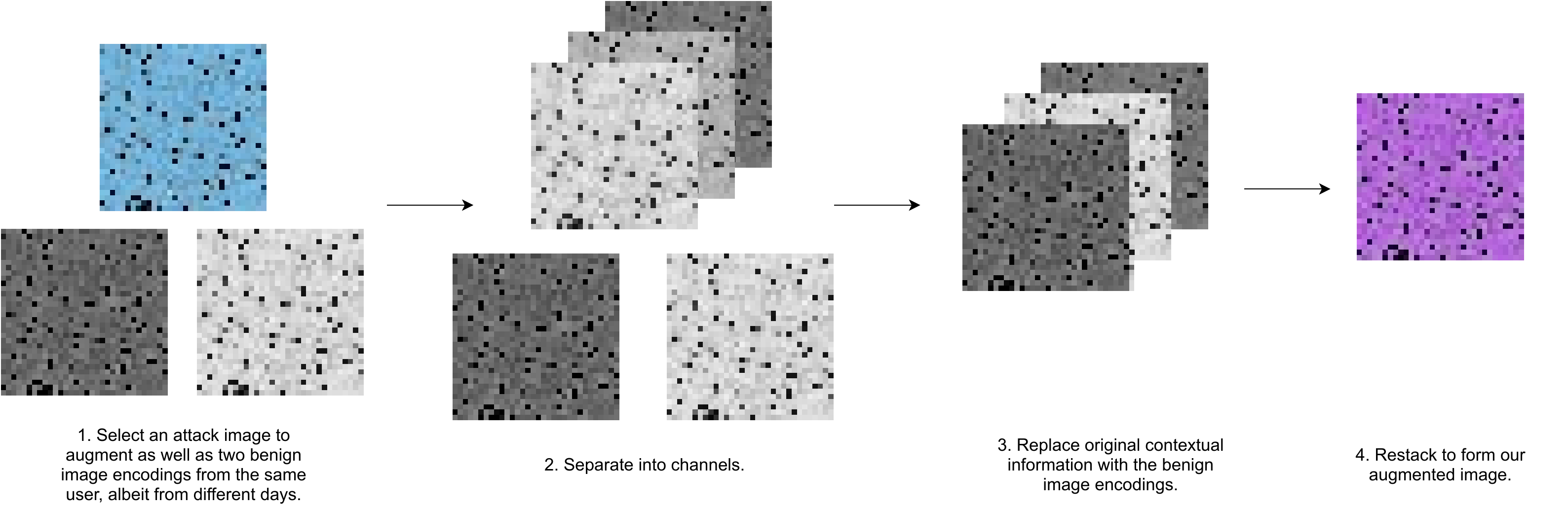}
   \caption{}
\end{subfigure}

\caption{\label{figure:augment} Context Changing Data Augmentations. (a) Channel swapping augmentation. (b) Channel replacement augmentation.}
\vspace{-5mm}
\end{figure*}

\subsection{\label{section:compTradEnc} Comparison to Traditional Encodings}
To better ascertain why this form of image encoding is superior to traditional encoding structures, it is important to understand how convolutional neural networks (CNNs) work. 

CNNs were developed as regular neural networks do not scale well for classification of images. For example, in datasets like CIFAR-10 \cite{krizhevsky2009learning} and ImageNet \cite{deng2009imagenet}, images are of size 32x32x3, so a single fully-connected neuron in a first hidden layer of a regular neural network would have 32*32*3 = 3072 weights. Typically for classification we would like for such networks to have several such neurons, and thus the number of parameters would grow unmanageable quickly. Full connectivity in networks designed for image classification is wasteful and the huge number of parameters would quickly lead to brittle models with poor generalization.

Thus, CNNs connect each neuron to only a local region of the input volume. The reasoning behind this decision is that if one feature is useful to compute at some spatial position (x,y), then it should also be useful to compute at a different position (x2,y2). As a result, CNNs classify images of various types by looking at spatial cues, where filters look for certain things in images and the location within the image does not matter. They are translation invariant and assume that certain patterns can occur in various parts of an image.

This assumption, known as the weight sharing assumption, does not hold for traditional image encodings. Not only are certain textures and patterns in an image encoding important to classification, so is the location of said features within the image; a CNN architecture would not be able to make these distinctions within a traditional image encoding.

In our novel color-based image encoding however, we do not need the CNN model to identify location-important features within an image. As malicious behavior is identifiable by cross-channel differences, the convolution filters within the CNN need to simply detect color anywhere within the image. As we can see in Figure \ref{figure:benignVsMalicious}, the distinguishable characteristics of malicious image encodings are translation invariant and satisfy the weight sharing assumption, leading to less brittle models that generalize better to unseen data.

\section{Context Changing Data Augmentation \label{section:dataAug}}

\textbf{Imbalance and Insider Threats:  } As the vast majority of employees within a company are good-natured and do not have malicious intent \cite{gheyas2016detection, azaria2014behavioral, sheykhkanloo2020insider, al2021integrated, gayathri2020image}, insider threat detection is a highly imbalanced problem space. This poses problems for potential classification models, as during training models will spend most of their time on the dominant class and will fail to learn enough from the minority classes; decision boundaries either become too complex and we lose the ability to generalize to unseen data, or minority sub-concepts are ignored altogether due to not providing enough discriminative information to classifiers \cite{fernandez2018learning}. The data imbalance problem is one of the most important unsolved challenges for current insider threat detection systems \cite{yuan2021deep}.

\textbf{Issues with Standard Approaches:  } Data augmentation is a popular technique used for a variety of different domains to help handle data imbalance issues \cite{afzal2019data, jiang2020data, ibrahim2018imbalanced, saini2020deep, shorten2019survey}. While basic image transformations such as cropping, rotations, and color shifting are computationally efficient methods that have been shown to work in theory \cite{shorten2019survey}, they are not suitable for our current image representations.

As our images are created via a SAE encoding, there are regions of sparseness that exist within each of the created images at the same locations. These regions of sparseness can be identified via the spots that occur in the greyscale images seen in Figures \ref{figure:greyToColor} and \ref{figure:benignVsMalicious}, as well as the final color images in Figures \ref{figure:benignVsMalicious} and \ref{figure:augment}. Since all evaluated data will have these spots in the same locations, augmentations such as cropping and rotations are not ideal as these will shift the locations of these spots in the augmented images, leading to our model learning representations that will not exist in true data. Similarly, our images are designed such that cross-channel differences are what enable us to define behavior as malicious; we cannot use color shifting as it will subvert the image representation and thus the predictive power of our classifier.

\textbf{Novel Solution:  } We instead observe that two channels of each image consist solely of information that allows us to identify if the current day's behavior is malicious. Thus, we are able to swap out information held in these channels and still possess a valid representation. We augment our training set by swapping the positions of the contextual channels; the channel corresponding to yesterday becomes the day before yesterday, and the channel corresponding to the day before yesterday becomes yesterday. Additionally, we also randomly select days of benign behavior for the given user and use these to replace our contextual information. Both forms of data augmentation we perform as well as examples of final results are illustrated in Figure \ref{figure:augment}.

\section{Attack Classification}

\begin{figure}[htbp]
\vspace{-4mm}
\centerline{\includegraphics[width=\linewidth,keepaspectratio]{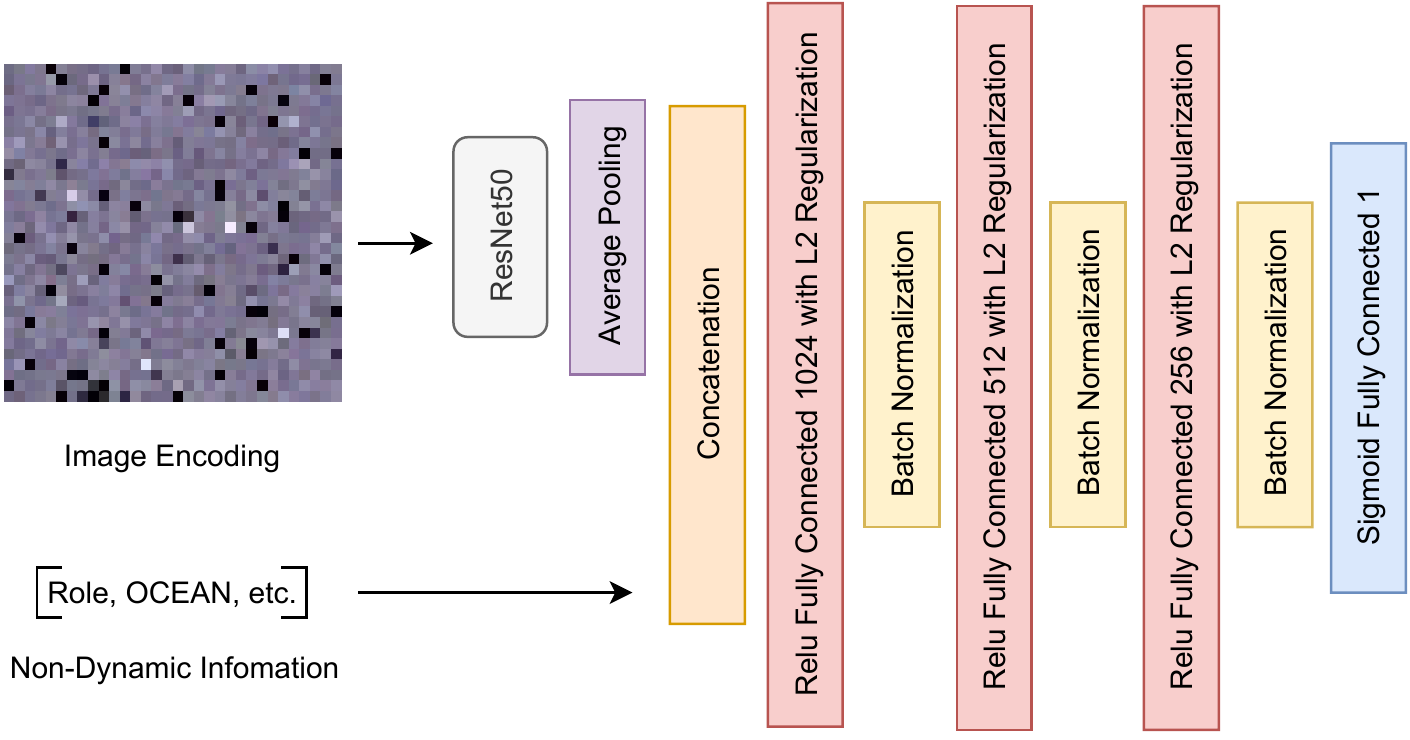}}
\caption{\label{figure:classArchitecture} Attack Classification Architecture. We pass in the encoded image as well as non-dynamic information such as role and psychometrics, and the proposed architecture outputs a classification of the given behavior as benign or malicious.}
\label{fig}
\vspace{-4mm}
\end{figure}

For reasons detailed in Section \ref{section:intro}, we elect to utilize transfer learning for our task, adopting the ResNet50 architecture \cite{he2016deep}. After training ResNet50 on the ImageNet dataset \cite{ILSVRC15}, we freeze the entire architecture except the last residual block, allowing for the final features extracted from the model to be tweaked during training.

While the image encoding enables us to identify malicious behavior at a glance, it is beneficial to provide additional information regarding a user, such as their role, to our classification model that could prove to be useful. Contextualizing the behavior of a user with background information can help to reduce the number of false positives a UEBA system reports. For example, an employee logging into systems late at night might be considered suspicious and by itself may lead to an alert being generated, but this activity should be considered normal for a security guard that is working a graveyard shift that given day.

In addition to role information, we also incorporate psychometric information, which can give us a glimpse into a person's psyche and determine if they are more likely to perform insider attacks. The disregard of such information is a crucial shortcoming of most insider threat detection regimes; approaches often neglect, or only make a superficial reference to, underlying psychological processes that might give rise to behavior that leads to insider attacks \cite{yuan2021deep, maasberg2015dark, sarkar2010assessing}. While most insiders do not sit with psychologists in situations where such information can be easily compiled, it is becoming more common over time for companies to require psychometric tests to be taken by potential candidates or incorporate psychometric properties into the hiring process \cite{interviewAICPA, mccarthy2004measuring, campion1988structured, gaurdianpsycho}; scoring metrics like OCEAN can easily be derived from such evaluations \cite{poropat2009meta}.

We concatenate these non-dynamic features with the output of the ResNet50 architecture, passing this vital information into fully connected layers before classification. Figure \ref{figure:classArchitecture} details the model architecture utilized.

\section{Experiments}
\textbf{Dataset:  } Traditional UEBA training data is composed of real-life scenarios, consisting of confidential information for a company, and the personal information of their employees. Thus, each vendor utilizes their own private datasets, making model comparisons and benchmarking difficult in nature. The CERT Insider Threat center together with ExactData LLC analyzed 1,154 actual insider incidents in the United States to create the largest public repository of insider threat scenarios in order to tackle this issue \cite{glasser2013bridging}. Many publications and companies have utilized this dataset to assess model architectures, perform integration testing, and run confirmatory hypothesis testing, solidifying its status as the gold standard public dataset for insider threat detection system benchmarking. \cite{lindauer2014generating}. 

The CERT Insider Threat dataset contains 32,770,224 unique events, with available audit data sources including logon activity, email traffic, web browsing traces, file access logs, thumb drive usage, as well as LDAP information describing the organization hierarchy and user roles.

Malicious users within the dataset execute activities in highly variable time periods, with some attacks completing within a day, while others occur over a 2 month span. The high diversity in attack time frames enables robust verification checks against insiders that intentionally act slowly.

While CVUEBA was initially designed and evaluated against a custom-built private dataset, for benchmarking and reporting purposes we will use version 4.2 of the CERT dataset as it is by far the most popular dataset and version combination used in insider threat publications.

\textbf{Attack Scenarios:  } There are three scenarios of attack within the dataset. In Scenario 1, a user obtains sensitive information they subsequently upload to Wikileaks. In Scenario 2, a user browses job sites looking for a job, stealing confidential information and leaving as soon as they find one. Finally, in Scenario 3 a system administrator grows to dislike their job and downloads and installs a keylogger onto their supervisor's computer. Using the obtained password, they send an alarming mass email acting as the supervisor, leaving the organization immediately afterwards. As occurs in the real world, this dataset is extremely imbalanced; Table \ref{table:imbalance} details the occurrence of attack scenarios compared to normal behavior within 24 hour time frames per user.

\begin{table}[htbp]
\caption{\label{table:imbalance} Imbalance Ratios Within Insider Threat Data}
\centering
\resizebox{0.7\linewidth}{!}{%
\begin{tabular}{@{}ccc@{}}
\toprule
Class Type & Number of Instances & Imbalance Ratio \\ \midrule
Normal     & 330452 & 1 : 1           \\
Scenario 1 & 85 & 1 : 3899        \\
Scenario 2 & 861 & 1 : 384         \\
Scenario 3 & 20 & 1 : 16570       \\ \bottomrule
\end{tabular}%
}
\end{table}

\textbf{Data Organization:  } Across experiments, data was categorized into different sets at the user level within 24-hour windows via a stratified split, with 70\% in the training set, 10\% in the validation set, and 20\% in the test set. CC was trained and tuned first using the training and validation sets, then the rest of the features were compiled afterwards. Insider threat systems typically classify behavior in a binary fashion as either malicious or benign behavior; we thus categorize all attack scenarios as malicious behavior.

\subsection{Relation Between Color and Behavior \label{subsection:coloreval}}
As discussed in Section \ref{subsection:contextchannel}, our image encodings are designed with the intent of using color to identify malicious behavior. In order to evaluate this design, we quantify the colorfulness of an image via the colourfulness metric \cite{hasler2003measuring}. As the category of behavior is binary, we utilize Point Biserial Correlation for evaluation \cite{tate1954correlation}. We compute the colorfulness metric for our image encodings and treat this as a continuous random variable, and we treat the true behavior classification as a binary random variable. For this experiment only, we classify an encoding as malicious if any of the channels correspond to malicious behavior in the ground truth; if the user acted maliciously previously, our image encoding should appear colorful in this scenario as well. To develop the image encodings, a Sparse AutoEncoder with one hidden layer of size 1024, using SeLU activations, a batch size of 220, a $\beta$ of 0.68, and $\rho$ of 0.45 is trained on the training set with a learning rate of 0.0001 using the NAdam optimizer \cite{tato2018improving}. To prevent regions of sparsity having a detrimental effect on the colorfulness metric for an encoding they are replaced by the mean pixel value of the given encoding.

\begin{table}[htbp]
\caption{\label{table:color} Correlation Between Color and Behavior}
\centering
\resizebox{0.35\linewidth}{!}{%
\begin{tabular}{@{}cc@{}}
\toprule
Representation & Correlation \\ \midrule
Daily     & \textbf{0.9301}           \\
Historical & 0.8884       \\
Role & 0.8342         \\\bottomrule
\end{tabular}%
}
\end{table}

Table \ref{table:color} shows the correlation results for each of the three context-channel representations detailed in Section \ref{subsection:contextchannel}. We obtain a strong correlation value of 0.9301 for our daily representations, indicating our behavior encoding properly represents malicious activity by color. Daily outperforming alternative representations is to be expected as this matches the underlying behavior observed in the CERT Insider Threat dataset where an attacker may act maliciously one day and act normally the next. Given these results, we utilize the daily representation for the experiments below.

\subsection{Model Evaluation \label{section:modeleval}}

\textbf{Baseline Comparison:  } Next, we evaluate CVUEBA performance in comparison against baseline models, comparing against VGG-19, MobileNetV2, and ResNet-50 models built in the same manner as the original paper \cite{gayathri2020image}, as well as the industry models detailed in Section \ref{section:relatedworks}.

Optimal hyperparameters for each industry model were identified by using Tree Parzen Estimation \cite{bergstra2013making}. The SAE network is trained in the manner detailed in Section \ref{subsection:greyscale} using the set of hyperparameters detailed in Section \ref{subsection:coloreval}. Our dual input network is trained using a batch size of 128, a learning rate of 0.01, ReLU activation functions, and uses the Adam optimizer \cite{kingma2014adam}.

Results can be found in Table \ref{table:baselinecomp}. We report balanced accuracy, precision, recall, and F1 score as these metrics allow for a good evaluation of models in problem spaces with high data imbalance \cite{fernandez2018learning}. 

For every computed metric, CVUEBA outperforms all baselines, with all reported metrics scoring above 98\%. 

\begin{table}[htbp]
\centering
\caption{\label{table:baselinecomp} Baselines vs CVUEBA}
\resizebox{\linewidth}{!}{%
\begin{tabular}{@{}cccccc@{}}
\toprule
Model & Balanced Accuracy & Precision & Recall  & F1 Score\\ \midrule
CVUEBA & \textbf{0.9923} & \textbf{0.9896} & \textbf{0.9845} & \textbf{0.9871} \\
Mahalanobis Distance & 0.5063 & 0.0154 & 0.0155 & 0.0154 \\
Naive Bayes & 0.8767 & 0.1028 & 0.7732 & 0.1815 \\
Support Vector Machine & 0.5345 & 0.0088 & 0.1031 & 0.0162 \\
VGG-19 & 0.9741 & 0.9293 & 0.9485 & 0.9388 \\
Logistic Regression & 0.9539 & 0.0640 & 0.9485 & 0.1198 \\
Factorization Machine & 0.8582 & 0.9456 & 0.7165 & 0.8152 \\
ResNet-50 & 0.9664 & 0.9141 & 0.9330 & 0.9235 \\
MobileNetV2 & 0.9664 & 0.9141 & 0.9330 & 0.9235 \\
\bottomrule
\end{tabular}%
}
\end{table}

\textbf{Comparison to State of the Art:  } Comparisons of CVUEBA to the best performing state of the art methods evaluated on the same benchmark can be found in Table \ref{table:sotacomp}. Various papers use different metrics for performance evaluation, however most report accuracy; for the sake of comparison, we do the same here. As can be seen, CVUEBA outperforms alternatives by atleast 3.61\%.

\begin{table}[htbp]
\caption{\label{table:sotacomp} CVUEBA vs Reported State of the Art Accuracy Values}
\centering
\resizebox{0.9\linewidth}{!}{%
\begin{tabular}{@{}ccc@{}}
\toprule
Method                           & Source                                                                                   & Accuracy \\ \midrule
CVUEBA                           & Proposed Approach                                                                        & \textbf{0.9995}   \\
Image Feature Representation & Gayathri et al. \cite{gayathri2020image}                                                     & 0.9634 \\ 
Boosted Logistic Regression      & Noever et al. \cite{noever2019classifier}                                                & 0.9600   \\
AD-DNN                       & Al-Mhiqani et al. \cite{al2021integrated}                                                    & 0.9600   \\
Graph Convolutional Networks & Zhou et al. \cite{zhou2005training}                                                          & 0.9450 \\
Random Forest with Randomization & Noever et al. \cite{noever2019classifier}                                                & 0.9400   \\
LSTM-RNN                     & Meng et al. \cite{meng2018deep}                                                              & 0.9385 \\
LSTM-Autoencoder             & Sharma et al. \cite{sharma2020user}                                                          & 0.9017 \\
Random Forest                    & Noever et al. \cite{noever2019classifier}                                                & 0.9000   \\
DBN-OCSVM                        & Lin et al. \cite{lin2017insider}                                                         & 0.8779   \\
DBN                              & Lin et al. \cite{lin2017insider}                                                         & 0.8442   \\
\bottomrule
\end{tabular}%
}
\end{table}

\subsection{Activation Maximization}

We have designed image encodings that allow for easy identification of malicious behavior and shown that the color of our encodings is strongly correlated to said behavior. We now seek to close the loop on encoding brittleness concerns by confirming that CVUEBA classifies behavior primarily by the color of the encoding. To this end, we elect to utilize Activation Maximization \cite{erhan2009visualizing}. Activation Maximization allows us to create visualizations that our model will perceive with the highest confidence as being either normal behavior or malicious behavior. These visualizations enable us to verify that our model is identifying the features of our generated images we wish for it to focus on and ignoring noise and other features we wish for it to disregard.

\begin{figure}[htbp]
\centerline{\includegraphics[width=\linewidth,keepaspectratio]{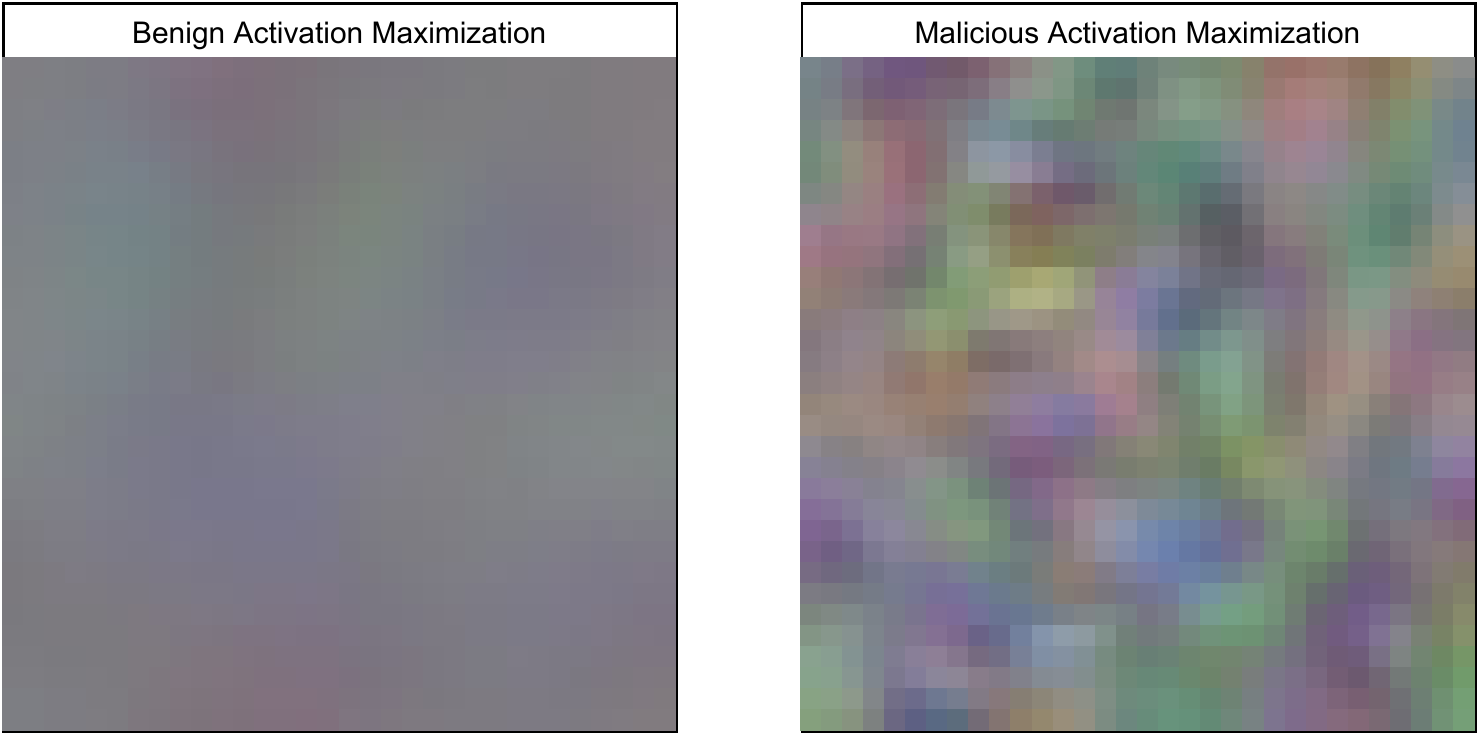}}
\caption{\label{figure:activationMaximization} Activation Maximization Images from the Attack Classifier}
\label{fig}
\end{figure}

Figure \ref{figure:activationMaximization} shows the resulting images from performing Activation Maximization for the benign and the malicious class. The malicious image is filled with various colors, while the benign image is a primarily grey image. This indicates that our model correctly focuses primarily on cross channel differences within the image as we expect.

We additionally note the absence of black spots on either image despite their presence on both benign and malicious images. Additionally, both images also show a lack of distinctive texture or information present in a specific location of the image. This indicates that our model correctly ignores the regions of sparsity and other sources of potentially specious information within the images, and validates that our image encodings correctly satisfy the weight sharing assumption.

\subsection{Evaluation on New Attack Types}
So far, CVUEBA has been trained, evaluated, and compared against baselines in scenarios where all models have seen similar attack scenarios before. As an insider threat detection system, it is imperative that a model can generalize well to identify attack scenarios without prior knowledge.

To evaluate this, we remove attack scenarios 1 and 3 from the training and validation sets. Scenario 2 was chosen to remain in the training set in order to minimize the likelihood of further data imbalance acting as a confounding variable due to its large class size in comparison to other the other attack scenarios in the dataset. All models are trained as detailed in Section \ref{section:modeleval}. Results are shown in Table \ref{table:unseen} and statistics regarding new attack vectors only can be found in Table \ref{table:unseenOnly}.

\begin{table}[htbp]
\centering
\caption{\label{table:unseen} Evaluation with Unseen Attack Vectors}
\resizebox{\linewidth}{!}{%
\begin{tabular}{@{}ccccccc@{}}
\toprule
Model & Balanced Accuracy & Precision & Recall  & F1 Score\\ \midrule
CVUEBA & \textbf{0.9871} & \textbf{0.9895} & \textbf{0.9742} & \textbf{0.9818} \\
Factorization Machine & 0.8376 & 0.9291 & 0.6753 & 0.7821 \\
Mahalanobis Distance & 0.5013 & 0.0059 & 0.0052 & 0.0055 \\
Naive Bayes & 0.8437 & 0.0991 & 0.7062 & 0.1739 \\
Logistic Regression & 0.8343 & 0.0699 & 0.6959 & 0.1270 \\
Support Vector Machine & 0.5253 & 0.0064 & 0.0928 & 0.0120 \\
VGG-19 & 0.9561 & 0.9124 & 0.9124 & 0.9124 \\
ResNet-50 & 0.9586 & 0.9036 & 0.9175 & 0.9105 \\
MobileNetV2 & 0.9535 & 0.8980 & 0.9072 & 0.9026 \\
\bottomrule
\end{tabular}%
}
\end{table}

\begin{table}[htbp]
\centering
\caption{\label{table:unseenOnly} Performance on New Attack Vectors Only}
\resizebox{\linewidth}{!}{%
\begin{tabular}{@{}ccccccc@{}}
\toprule
Model & Number of Attacks Detected & Percent Classified Correctly\\ \midrule
CVUEBA & \textbf{101} & \textbf{0.9619}\\
Naive Bayes & 88 & 0.8380\\
Factorization Machine & 87 & 0.8285\\
Logistic Regression & 86 & 0.8190\\
ResNet-50 & 85 & 0.8095\\
VGG-19 & 82 & 0.7809\\
MobileNetV2 & 79 & 0.7523\\
Support Vector Machine & 54 & 0.5142\\
Mahalanobis Distance & 52 & 0.4952\\

\bottomrule
\end{tabular}%
}
\end{table}

CVUBA performs only slightly worse when introduced to new attack vectors, unlike the baseline models, which perform much worse. This is due to CVUEBA detecting behavioral shifts rather than specific patterns of attack. The Computer Vision baselines using traditional image encodings have far worse detection performance on new attacks when compared attacks seen beforehand, indicating that these models are overfitting to the data and are unable to properly learn adequate information from these spatial representations to make informed decisions regarding the behavior of employees. CVUEBA generalizes better, is less brittle, and is more appropriate for deployment in a real-world setting.

\subsection{Ablation Studies}

In modern machine learning, black box models are widely used due to their high performance, however in cybersecurity it is important to develop and deploy models that are easily understood. Here we ablate important design elements of CVUEBA to achieve this \cite{reddy1975speech}.

\textbf{Architecture Evaluation:  } Table \ref{table:ablation} showcases the results for ablation studies designed to evaluate various architecture decisions. Models are trained as detailed in Section \ref{section:modeleval}, and we go into detail regarding the reasoning behind the ablation choices below.

\begin{table}[htbp]
\centering
\caption{Architecture Study \label{table:ablation}}
\resizebox{\linewidth}{!}{%
\begin{tabular}{@{}ccccccc@{}}
\toprule
Modification & Balanced Accuracy & Precision & Recall  & F1 Score\\ \midrule
None & 0.9923 & \textbf{0.9896} & 0.9845 & \textbf{0.9871} \\
Image only & \textbf{0.9947} & 0.8889 & \textbf{0.9897} & 0.9366 \\
Used VGG-19 & 0.9819 & 0.9590 & 0.9639 & 0.9614 \\
Used MobileNetV2 & 0.9793 & 0.9637 & 0.9588 & 0.9612 \\
Used Greyscale & 0.9690 & 0.9733 & 0.9381 & 0.9554 \\
Interpolation & 0.9896 & 0.9694 & 0.9794 & 0.9744 \\
Fully Connected & 0.9484 & 0.9305 & 0.8969 & 0.9134 \\
Non-dynamic only & 0.5827 & 0.0263 & 0.1856 & 0.0460 \\
\bottomrule
\end{tabular}%
}
\end{table}

For the image only model, we evaluate how well our image encodings by themselves perform in identifying malicious behavior by removing the concatenation layer as well as the contextual input from the model. The architecture used for this model is similar in form and function as what is used by our ResNet baseline and thus allows us to directly compare our proposed image encodings to the interpolation procedure used by the Image Feature Representation model. 

For the non-dynamic feature only model, we evaluate how well features such as role and psychometrics by themselves can be used to predict an attack by removing the concatenation layer and ResNet50 layers from CVUEBA.

Next, we wish to determine the benefit of using an image encoding and natural image pretraining in the first place. To this end, we trained a Feed-Fordward Fully Connected Neural Network architecture with ReLU activations and 6 hidden layers of sizes 2048, 2048, 1024, 1024, 512, and 256, with Batch Normalization layers in-between each Dense layer \cite{ioffe2015batch}, and a sigmoid output. Rather than use a ResNet architecture, our behavior encoding images are flattened and concatenated immediately with contextual information. Note this Fully Connected model maintains the other benefits of CVUEBA such as the introduction of non-dynamic information and SAE feature projection.

We use a ResNet50 architecture as part of CVUEBA. As the CNN is an important part of the CVUEBA architecture, we wish to evaluate what performance differences if we used alternative CNN models like a VGG-19, or MobileNetV2 architecture instead.

Additionally, we utilize context channels to develop color image encodings. We seek to determine how it would affect the performance of CVUEBA if we instead used the original greyscale encodings developed in Section \ref{subsection:greyscale}. While this image encoding implementation does benefit from the SAE model projecting malicious behavior away from benign behavior, this image encoding implementation is still privy to the issues with traditional image encodings as detailed in Section \ref{section:compTradEnc}.

In addition to the novel image encoding proposed, CVUEBA has additional features that lead to its superior performance, such as the dual-input model architecture. For a more direct comparison to traditional encodings, we utilize the interpolation based image encoding procedure utilized by previous research on natural image transfer learning for insider threat detection \cite{gayathri2020image}. 

All Computer Vision approaches outperform our Fully Connected architecture, indicating that using natural image transfer learning does improve model performance. When we compare the interpolation model with CVUEBA, and the image only model to the ResNet baseline, we see that our image encodings lead to superior results on similar model architectures. We also find that the color image encoding models perform significantly better in terms of recall and precision when compared to models trained on the initial greyscale image encodings. When comparing CVUEBA to the image only model, we see that adding the non-dynamic information reduces recall slightly, but leads to a huge improvement in precision. As false positives could lead to an employee losing his/her job, this trade-off is beneficial. As one may expect, the non-dynamic architecture performed the worse, however its performance was better than random guessing, providing additional credence to the use of these features in the CVUEBA architecture.

\textbf{Augmentation Evaluation:  } For augmentation ablations, we wish to study various augmentations in relation to CVUEBA. Results are shown in Table \ref{table:augmentation}. The context changing augmentation outperforms all alternatives, and color shift leads to performance degradation. Random flips, rotations, and cropping would lead to features of the image encodings being shifted location-wise. These augmentations performing better than using no augmentations provides further credence to our color-based augmentations fulfilling the weight sharing assumption. The context changing augmentation outperforms all alternatives supports the theory detailed in Section \ref{section:dataAug}.

\begin{table}[htbp]
\centering
\caption{Augmentation Study \label{table:augmentation}}
\resizebox{\linewidth}{!}{%
\begin{tabular}{@{}ccccccc@{}}
\toprule
Augmentation & Balanced Accuracy & Precision & Recall  & F1 Score \\ \midrule
Context Changing & \textbf{0.9923} & \textbf{0.9896} & \textbf{0.9845} & \textbf{0.9871} \\
Random Flip & 0.9742 & 0.9534 & 0.9485 & 0.9509 \\
Random Rotation & 0.9716 & 0.9581 & 0.9433 & 0.9506 \\
Random Crop & 0.9690 & 0.9381 & 0.9381 & 0.9381 \\
None & 0.9638 & 0.9326 & 0.9278 & 0.9302 \\
Color Shift & 0.9246 & 0.6548 & 0.8505 & 0.7399 \\
\bottomrule
\end{tabular}%
}
\end{table}

\section{Conclusion}
CVUEBA takes the complex problem of insider threat detection and simplifies it down to the simpler problem of color detection, using the behavioral changes that occur during attacks to its advantage by representing contextual information via different channels. We show that this novel image encoding procedure leads to models that have greater predictive power when evaluated on attack vectors similar to those seen in training data while also generalizing far better to detect new forms of attacks as well. An advanced feature set, a powerful color-based encoding structure, a novel method of tackling imbalanced learning, and a dual input classifier help CVUEBA outperform state of the art models in both academia as well as industry.

\section*{Availability}
Sample code and data can be found here: \url{https://github.com/CVUEBARepo/CVUEBA}

\printbibliography

\appendix

\section{Quantifying Importance of Proposed Features}

In cybersecurity it is crucial for machine learning models and predictions to be as transparent as possible. An important method of approaching this is issue is from a causality frame of mind where we aim to determine what makes the detection model behave in a certain way. As each feature is a potential cause behind the model labeling an employee's behavior as malicious it is important to quantify the degree of influence each feature has on the prediction's made by the model.

In the main manuscript, we propose novel features based on calculating the FPV of accessed files and extracting novel indicators from text using CC. Here, we wish to evaluate the importance of these features to classification performance using a feature removal paradigm. To this end, we utilize the Explaining by Removing framework (ERF) \cite{covert2021explaining}, which assesses and unifies 26 existing feature removal methods.  The framework simplifies the feature removal task into the making of three choices: how it removes features, what model behavior it analyzes, and how it summarizes feature influence. 

As suggested by ERF, we elect to remove features by marginalizing them out using their conditional distributions $p \left( X_S \mid X_S = x_s \right)$. This approach is shown in Equation 8. Here, F denotes a surrogate model trained to imitate the original model's prediction while taking in a subset of the given features denoted as $S$. $X_S$ refers to the restricted input domain after applying the subset, and $x_S$ is an instance in the domain.

\begin{equation}
    F(x_S) = \mathop{{}\mathbb{E}} \left[ f(X) \mid X_S = x_s \right]
\end{equation}

Marginalizing out using conditional distributions is quantifies how much information is provided by knowing a feature's value. It has a strong foundation in information theory, and strong ties to a variety of model explainability methods \cite{lundberg2017unified, lundberg2020local, covert2020understanding}, making it an ideal choice for feature removal.

In order to evaluate the importance of the proposed features in detecting all scenarios of attack, for the model behavior choice we assess the dataset loss as a whole. 

Table \ref{table:featcodenames} lists the feature codenames we are assessing, and Table \ref{table:featremoval} details the results of running feature removal. We report all summary techniques in ERF. All proposed features have a strong influence on the predictions made by CVUEBA.

\begin{table}[htbp]
\footnotesize
\caption{\label{table:featcodenames} Set of features evaluated}
\begin{tabularx}{\linewidth}{c*{1}{>{\raggedright\arraybackslash}X}}
\toprule
Codename & \multicolumn{1}{c}{Feature}                                                                      \\\toprule
FPV         & File Path Variance throughout the day.                                                       \\ \midrule
FPV\_After          & File Path Variance after office hours.                                                       \\ \midrule
CC\_Disgruntled         & Number of emails Conical Classification identified as the user being disgruntled.            \\ \midrule
CC\_Job          & Number of websites Conical Classification identified as being job posting sites.                     \\ \midrule
CC\_Wikileaks         & Number of websites Conical Classification identified as being Wikileaks or Wikileaks clones. \\ \midrule
CC\_Keylogger          & Number of websites Conical Classification identified as being keylogger download sites.      \\
\bottomrule
\end{tabularx}
\end{table}

\begin{table}[htbp]
\centering
\caption{\label{table:featremoval} Feature Removal Model Explanation}
\resizebox{\linewidth}{!}{%
\begin{tabular}{@{}cccccc@{}}
\toprule
Feature         & \begin{tabular}[c]{@{}c@{}}Shapely \\ Value\end{tabular} & \begin{tabular}[c]{@{}c@{}}Banzhaf \\ Value\end{tabular} & \begin{tabular}[c]{@{}c@{}}Remove\\ Individual\end{tabular} & \begin{tabular}[c]{@{}c@{}}Include\\ Individual\end{tabular} & \begin{tabular}[c]{@{}c@{}}Mean When\\ Included\end{tabular} \\ \midrule
FPV & 0.0427 & 0.0094 & 0.0096 & 0.0866 & -0.2551 \\
FPV\_After & 0.0022 & 0.0004 & 0.0004 & 0.0134 & -0.2760 \\
CC\_Disgruntled & 0.0021 & 0.0029 & 0.0032 & 0.0015 & -0.2764 \\
CC\_Job & 0.0050 & 0.0046 & 0.0046 & 0.0057 & -0.2758 \\
CC\_Wikileaks & 0.0584 & 0.0608 & 0.0199 & 0.0962 & -0.2490 \\
CC\_Keylogger & 0.0030 & 0.0040 & 0.0045 & 0.0014 & -0.2762 \\
\bottomrule
\end{tabular}%
}
\end{table}

\section{Deployment in Industry}

It is important to consider how an insider threat detection system will change, adapt, and improve when deployed in an industry setting; a static model is one that can be easily circumvented by malicious actors. We handle this by obtaining feedback from security experts in two stages: first when a new insider threat attack vector has been introduced, and second when our model identifies behavior as potentially malicious.

\subsection{Process of Handling New Attack Vectors}

The landscape of insider threats is ever changing, with new forms of attack found over time such as insider collusion attacks that have been growing rapidly in prevalence \cite{wang2020insider}. It is vital for our systems to adapt to these changes.
be able to grow and adapt to this ever-changing climate.

\begin{figure}[htbp]
\centerline{\includegraphics[width=\linewidth,keepaspectratio]{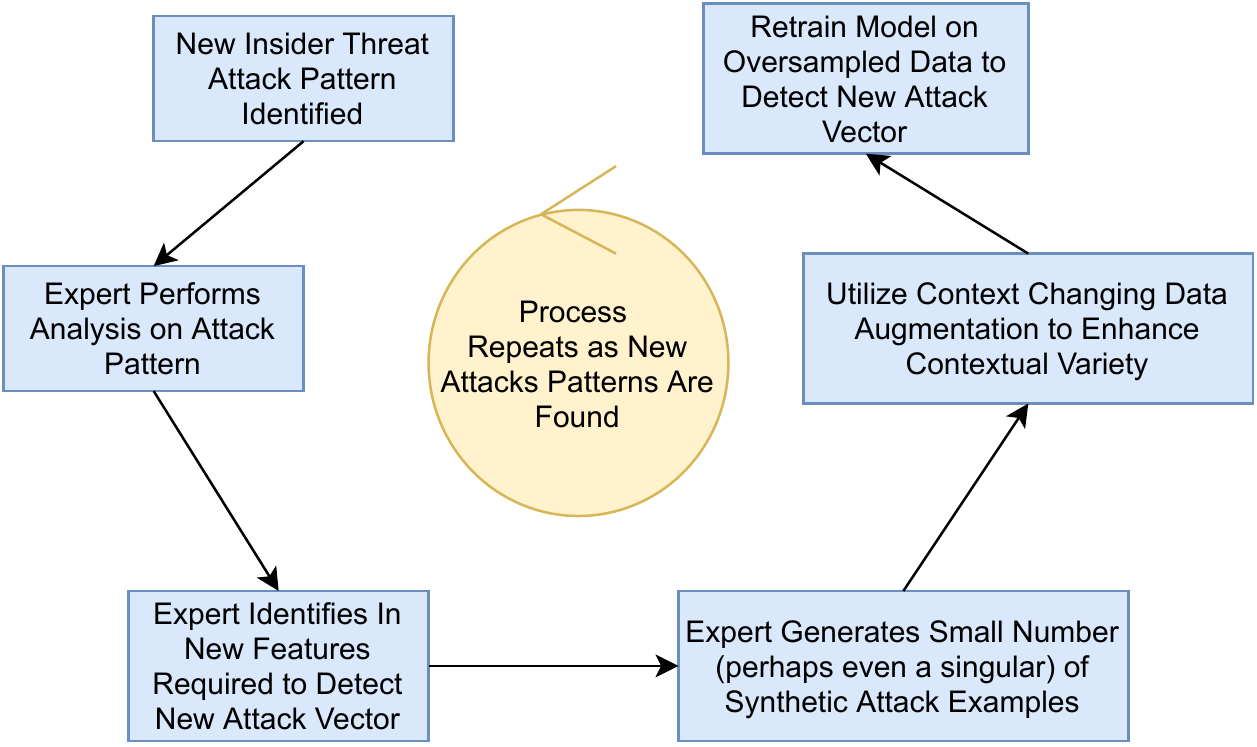}}
\caption{\label{figure:cycleattack} Cycle of incorporating new attack patterns.}
\label{fig}
\end{figure}

Figure \ref{figure:cycleattack} illustrates how an expert can update CVUEBA in accordance with the discovery of new methods of attack. If the current implementation is deemed to be outmoded after analysis of the attack vector and feature engineering is complete, we utilize the augmentation strategy proposed in the main to update CVUEBA, enabling security teams to detect and identify this new method of attack.

\begin{figure}[htbp]
\centerline{\includegraphics[width=\linewidth,keepaspectratio]{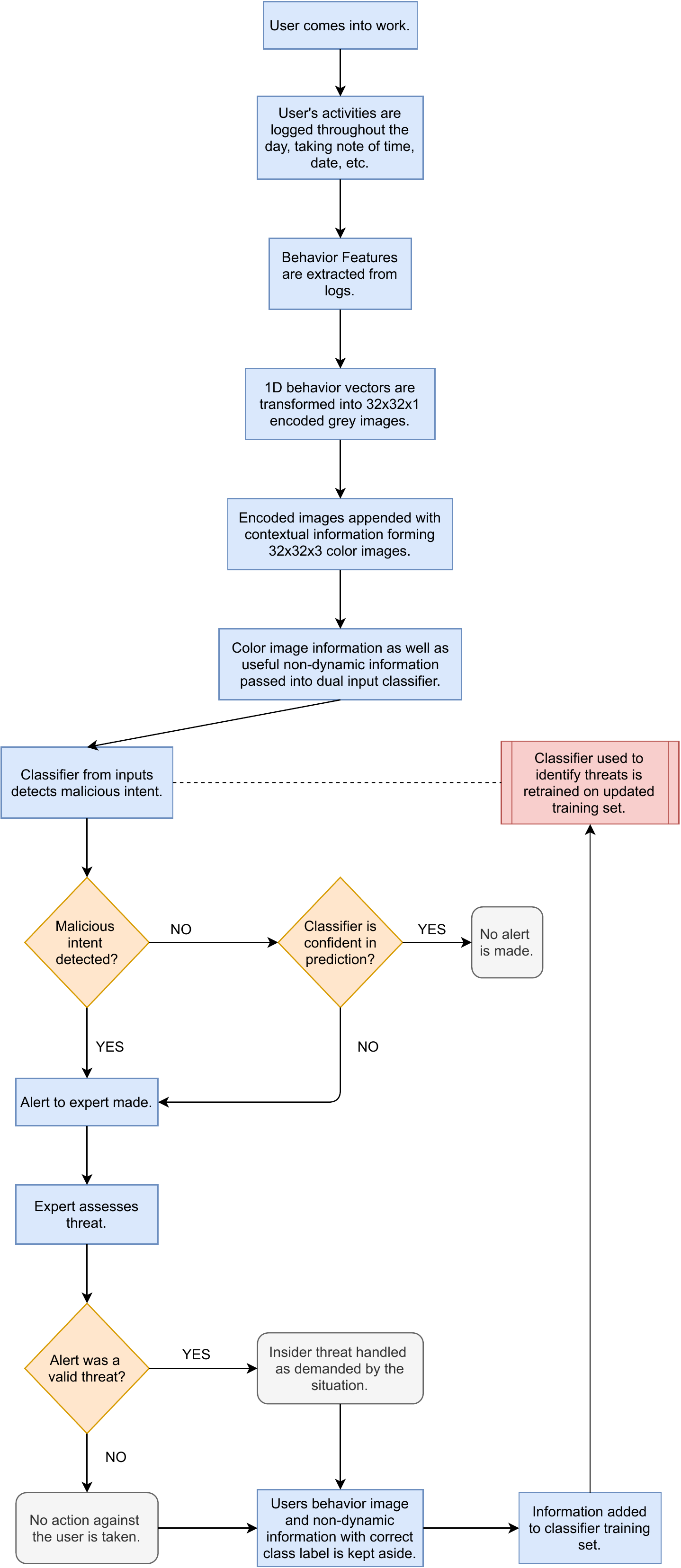}}
\caption{\label{figure:CVUEBAProcessFlow} Process of assessing and handling user behavior in a deployment setting.}
\label{fig}
\end{figure}

\subsection{Looping in the expert at detection time}

While CVUEBA has a false positive rate lower than alternative algorithms, false positives are still a serious issue for UEBA systems as a false positive could lead to a well-intentioned employee getting fired for no fault of their own.

Thus, we propose a method that allows security experts to catch false positives before action is taken, and allows for CVUEBA to adapt and improve based on expert feedback.

Figure \ref{figure:CVUEBAProcessFlow} illustrates how CVUEBA can improve and adapt to the expert's preferences after deployment. User information is passed through the model in 24 hour intervals every time a new batch of behavior is identified; this enables real-time detection of attacks as they occur.

As shown in the main manuscript, the model's output layer is a fully connected sigmoid; this enables us the model's confidence in the given classification by determining how close the output value is to 0.5. Based on a threshold set by the security team, the model sends alerts not only for behavior deemed malicious, but also behavior classified with low confidence.

The expert is then able to assess the alert from the image encodings created via the process detailed in the main manuscript as well as from the raw feature vectors detailed previously. If the user's behavior is deemed to be malicious, the insider threat is handled based on the given situation.

Regardless of the final conclusion, we can take advantage of the expert's inquiry to improve our model's performance. Taking advantage of Active Learning techniques \cite{settles2009active} with our expert performing the role of an oracle, the CVUEBA classifier is trained using the newly labeled vectors.

This enables our model to adapt to the ever-changing environment found in corporations, as well as to the preferences of the expert in charge of the CVUEBA deployment.

We seek to evaluate loop-in by initially training CVUEBA on the training set. Using a threshold parameter of 0.4, we simulate incoming data in a production environment via the validation set and simulate expert feedback using the ground truth labels. The improved model is evaluated on the test set in the same manner as the original model. Table \ref{table:loopin} showcases the results.

\begin{table}[htbp]
\centering
\caption{\label{table:loopin} Effects of Loop-in \vspace{1mm}}
\resizebox{\linewidth}{!}{%
\begin{tabular}{@{}cccccc@{}}
\toprule
With Loop-in & Balanced Accuracy & Precision & Recall  & F1 Score\\ \midrule
No & 0.9923 & 0.9896 & 0.9845 & 0.9871 \\
Yes & 0.9948 & 0.9948 & 0.9897 & 0.9922 \\
\bottomrule
\end{tabular}%
}
\end{table}

Loop-in offers a modest improvement in both precision and recall, leading to a further reduction in false positive rate as was intended by the design.

\section{Computer Vision Baseline Architectures}

\begin{figure}[htbp]
\centerline{\includegraphics[width=\linewidth,keepaspectratio]{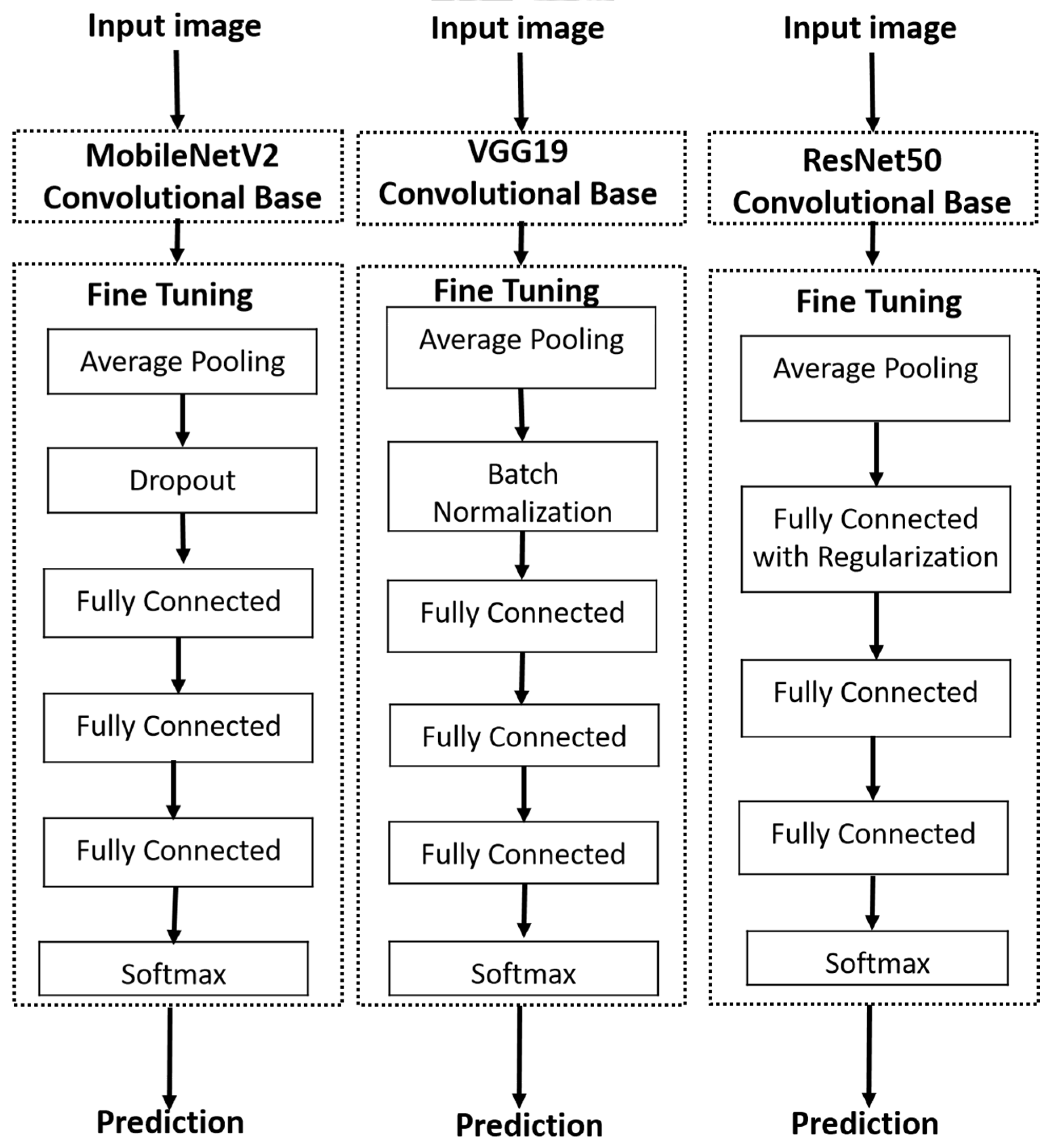}}
\caption{\label{figure:BaselineArchitecture} Computer Vision Baseline Architectures. These are the architectures used for evaluation purposes of CVUEBA, as defined and illustrated in the original paper \cite{gayathri2020image}.}
\label{fig}
\end{figure}

Figure \ref{figure:BaselineArchitecture} illustrates the architecture for each of the Computer Vision Models that serve as our main baselines for evaluation purposes.

The MobileNetV2 model is trained using RMSProp for 15 epochs using a batch size of 64, and a dropout rate of 0.3. The VGG19 model is trained using SGD for 15 epochs using a batch size of 128. Finally, the ResNet50 model is trained using Adamax for 15 epochs using a batch size of 128, and has an L2 regularization term. As the original paper does not specify the value of $\lambda$, we utilize the default value of $0.01$ defined by Tensorflow \cite{tensorflow2015-whitepaper}. All three models are trained using a undersampled version of the training set where the benign to malicious ratio is reduced to 20:1.

\end{document}